\documentclass[letterpaper, 10 pt, conference]{ieeeconf}

\IEEEoverridecommandlockouts                              

\usepackage{graphicx}
\usepackage{bm}
\usepackage[T1]{fontenc}
\usepackage{amsmath}
\usepackage{amsfonts}
\usepackage[colorlinks,
linkcolor=blue,
anchorcolor=blue,
citecolor=green
]{hyperref}
\usepackage{booktabs}
\usepackage{threeparttable}

\usepackage{color}   
\definecolor{g}{rgb}{0,0,0}
\usepackage{ulem}

\begin{document}
\title{\LARGE \bf
Enhancing Adaptivity of Two-Fingered Object Reorientation Using Tactile-based Online Optimization of Deconstructed Actions
}

\author{Qiyin Huang, Tiemin Li, and Yao Jiang
\thanks{This work was supported in part by the National Natural Science Foundation of China under Grant 52375017; in part by the National Natural Science Foundation of China under Grant 52175017; and in part by the Joint Fund of Advanced Aerospace Manufacturing Technology Research under Grant U2017202.}
\thanks{Qiyin Huang, Tiemin Li, and Yao Jiang are with the Institute of Manufacturing Engineering, Department of Mechanical Engineering,
Tsinghua University, Beijing 100084, China.
        {(e-mail: willhqy525@gmail.com; litm@mail.tsinghua.edu.cn; jiangyaonju@126.com)}}%
}

\maketitle
\thispagestyle{empty}
\pagestyle{empty}

\begin{abstract}
Object reorientation is a critical task for robotic grippers, especially when manipulating objects within constrained environments. The task poses significant challenges for motion planning due to the high-dimensional output actions with the complex input information, including unknown object properties and nonlinear contact forces. Traditional approaches simplify the problem by reducing degrees of freedom, limiting contact forms, or acquiring environment/object information in advance—significantly compromising adaptability. To address these challenges, we deconstruct the complex output actions into three fundamental types based on tactile sensing: task-oriented actions, constraint-oriented actions, and coordinating actions. These actions are then optimized online using gradient optimization to enhance adaptability. Key contributions include simplifying contact state perception, decomposing complex gripper actions, and enabling online action optimization for handling unknown objects or environmental constraints. Experimental results demonstrate that the proposed method is effective across a range of everyday objects, regardless of environmental contact. Additionally, the method exhibits robust performance even in the presence of unknown contacts and nonlinear external disturbances.

\end{abstract}
\section{INTRODUCTION}

Object reorientation is a fundamental task that highlights the dexterity of the gripper. \textcolor{g}{For example, when extracting an object from a container with a narrow opening, the robot may need to adjust the object's orientation within the gripper to ensure smooth removal. In such cases, reorientation becomes essential, as the initial grasp may not be ideal.} During this process, the controller must account for forces such as gravity or environmental contact, potentially using these forces to assist in adjusting the object's pose.

If object reorientation planning is viewed as a system, it presents significant challenges to motion planning both in terms of output and input. On the output side, the system's degrees of freedom encompass not only the gripper's spatial movement but also its gripping force, resulting in high-dimensional outputs that increase planning complexity. On the input side, the contact dynamics between the object and the environment are unknown and vary nonlinearly \cite{xianyi2024Enhancing}. Planning robot motion while considering these complex and dynamic inputs is especially challenging.

To address the aforementioned challenges, we propose a multi-action superposition control framework based on tactile proprioception to achieve pivoting of unknown objects. To simplify the control of high degrees of freedom, the output is divided into three fundamental actions: (1) Task-oriented actions aimed at completing the pivot rotation; (2) Constraint-based actions designed to prevent macroscopic tangential slip; (3) Coordinating actions intended to resolve conflicts between the first two actions. To streamline the planning process, the objectives of the three actions are based on simplified tactile expectations. This approach eliminates the need for real-time measurements of complex, nonlinear contact information. Finally, these actions are superimposed and key actions are optimized online using a gradient-based planning method, further enhancing the system's adaptability.

The contributions of this method are as follows: (1) Reformulation of planning objectives: reconstructing the complex environmental contact planning by focusing on the expectation of finger tactile sensations, and simplifying the task into a constrained planning problem; (2) Decomposition of complex actions: Breaking down the gripper's complex degrees of freedom into three directly superimposable actions based on task objectives; (3) Online optimization of actions: Utilizing an online gradient planning method that enables the robot to optimize its actions during the pivoting of unknown objects in real time, without requiring prior experience.

\section{Related Work}
\subsection{Tactile-based Reorientation}

Object reorientation, adjusting an object's pose relative to the gripper, is essential in tasks like tool manipulation and assembly \cite{SUOMALAINEN2022104224}, \cite{lee2022peg}. While multi-fingered hands can achieve reorientation of novel objects \cite{tao2023Visual}, \cite{openai:marcin2020Learning}, it remains challenging for two-fingered robotic grippers, which are widely used in academia and industry \cite{baohua2020Stateart}. However, studies show that even with simple contacts, complex reorientation tasks can be achieved through environmental interactions \cite{neel2022Manipulation}, \cite{yu2021Cable}. Exploring the capabilities of two-fingered grippers not only broadens their application but also enhances the dexterity of multi-fingered hands using just two fingers.

Two-fingered object reorientation can be classified into two types: without environmental contact and with environmental contact. For reorientation without contact, research focuses on analyzing contact states between the fingers and object to control the gripper \cite{simon2018Extrinsic}. These states can be obtained through model-based analysis \cite{marco2018Slipping}, learning \cite{shaowei2024LearningBased}, or real-time measurements \cite{ruomin2023Novel}. These states are generally classified into macroscopic tangential sliding and rotational sliding \cite{andrea2017Control}. The control of the gripper, particularly its grasping force, is then based on the contact state \cite{chao2021Status}. Tactile sensor advancements \cite{wei2018Tactile}, \cite{mingxuan2022Continuous} have enriched contact information, boosting research in this area \cite{jason2022InHand}, \cite{marco2020TwoFingered}. However, assumptions like planar objects or known friction coefficients are often made, and these studies usually neglect gripper pose adjustments, limiting the dexterity of the object reorientation.

The second category involves scenarios where the object interacts with the environment, introducing more complex dynamics. Early studies used open-loop control, allowing passive compliance with external contact \cite{anne2015General}. Modeling and planning of contacts is another approach for controlling the gripper \cite{nikhil2020Planar}. With tactile sensing, some studies have used real-time measurements of the object's shape \cite{maria2024SimPLE} and contact force \cite{sangwoon2023Simultaneous} to guide gripper actions \cite{miquel2024TactileDriven}, \cite{sangwoon2024TEXterity}. However, modeling these interactions is challenging due to their non-linear, discontinuous nature \cite{xianyi2024Enhancing}. Many studies simplify them, such as treating contacts as points, but this limits the methods' generalizability.

It is evident that existing methods impose strict constraints on the type of objects, contact forms, and degrees of freedom in gripper movements, significantly limiting the adaptability of these methods to different objects and reducing the dexterity of reorientation tasks. In this study, we aim to achieve more dexterous in-hand reorientation by leveraging tactile sensing while considering more general unknown objects and contact states, as well as incorporating degrees of freedom control, including distance between fingers and the posture of the gripper.

\subsection{Online Optimization for Actions}
During object reorientation, unpredictable factors like the object's properties, environmental contacts, gravity, and external disturbances can render pre-set control algorithms ineffective. Thus, online optimization is needed to allow real-time action updates for adaptability.

Popular online optimization methods in robot manipulation are generally divided into three categories. The first involves pre-training in a simulated environment followed by fine-tuning online \cite{gordon2020adaptive}. This approach requires extensive simulation and heavily depends on the quality of training data for new scenarios \cite{zhao2023skill}. Since this study aims for successful reorientation of unseen objects on the first attempt without relying on simulation for data collection, this method is not applicable. The second is Model Predictive Control \cite{francoisr2020Reactivea}. However, its reliance on precise system modeling, high computational overhead, and multi-step predictions makes it less ideal for object reorientation tasks, which often require real-time adaptability, reactive control, and learning from immediate feedback. The third method is online gradient optimization \cite{caporali2024deformable}, which continuously updates decision variables by computing or approximating gradients, making it ideal for continuous action spaces. Given the dynamic and unpredictable nature of object reorientation, this method is employed in this study. Before applying this method, it is necessary to define the action forms and design an appropriate loss function.

\begin{figure*}[t]
        \centering
        \includegraphics[width = 0.9\linewidth]{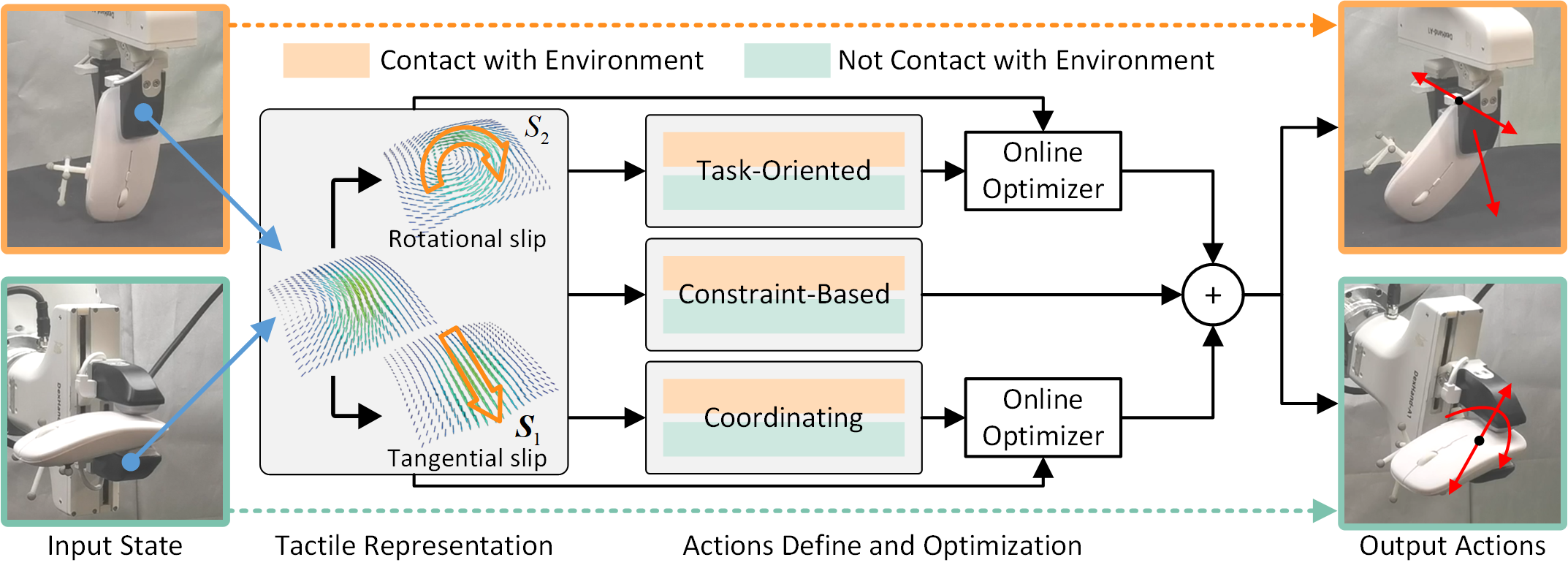}
        \caption{ \textcolor{g}{The orange and green arrows represent the reorientation tasks for the contact and non-contact states, respectively. Both states follow a common method, initiated by the blue arrows. The tactile information is represent as $\boldsymbol{S}_1$ and $S_2$, which are used to control the actions.} Of the three defined action types, task-oriented actions (when in contact with the environment) and coordinating actions (without environmental contact) do not require tactile feedback. These actions are optimized and learned online. Finally, different actions are superimposed to achieve the reorientation of the object.}
        \label{fig::framework}
\end{figure*}

\section{Method}
\subsection{Framework}

The purpose of this paper is to change the pose of an object relative to the gripper using pivot rotation, while ensuring that the object does not slip from the gripper during the process. Here pivot rotation means the object rotates relative to the gripper, and the axis of rotation passes through the contact area between the two fingers and the objects.  

The proposed controller is capable for the following two scenarios: (1) Contact state: the object is in contact with the environment. (2) Non-contact state: the object has no contact with the environment.

\textcolor{g}{The system's degrees of freedom consist of (1) the opening and closing of the gripper, which controls gripping force and (2) changes in the gripper's posture, including three-dimensional rotation and translation.} This work focuses on the control of the gripper. To better isolate the problem, we assume that the controller is provided with information about the contact scenario prior to initiating reorientation task. The framework is shown in Fig. \ref{fig::framework}. 

\subsection{Action Define}

\color{g}
The aim of this section is to determine the suitable form for each sub-action. In this context, "suitable" means the action does not need to be optimal, but effective enough to allow the robot to accomplish the intended goal.
\color{black}
For the reorientation task, two key golas must be satisfied: (1) Object rotation target: the object must rotate around the axis formed by the line connecting the finger contact surfaces; (2) Contact constraint condition: limit macroscopic sliding between the object and the fingers. Based on these goals, three types of actions are defined: 
\begin{itemize}
        \item Task-oriented action: complete the task (rotate the object).
        \item Constraint-based action: avoid undesired contact states (macro-tangential slippage of fingers and objects).
        \item coordinating action: Alleviate the contradiction between the above two actions.
\end{itemize}

\begin{figure}[t]
        \centering
        \includegraphics[width = 0.9\linewidth]{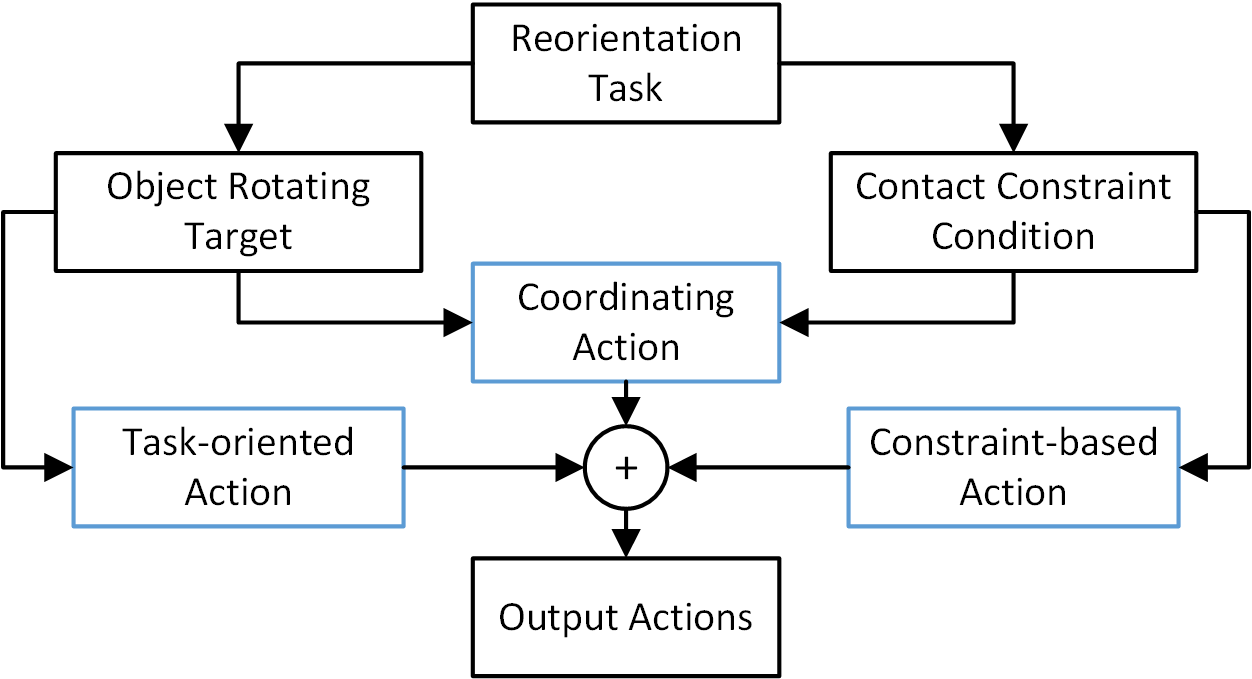}
        \caption{The reorientation task is divided into the object rotating target and the contact constraint condition. The former is achieved by task-based action, and the latter is achieved by constraint-based action.}
        \label{fig::actions frame}
\end{figure}

The relationship between different actions and their respective goals is illustrated in Fig. \ref{fig::actions frame}. For example, when extracting a slender object from a narrow opening, the gripper first adjusts the object's orientation by rotating it against the environment (task-oriented action). To prevent slippage, the gripping force may be increased (constraint-based action). If environmental contact causes unintended sliding, the gripper adjusts its pose (coordinating action).

The execution, direction, or amplitude of the actions is influenced by tactile feedback. To reduce computational complexity, the tactile input is simplified into two categories of trends: tangential slip and rotational tendencies. These trends are approximated using the following calculation method:
\begin{align}
        \label{equ::average tangential slip}
        \boldsymbol{S}_1&=1/\text{N} \sum\nolimits_{i=1}^N {}^{\mathbb{C}}\boldsymbol{d}_{i}, \\
        S_2&= 1/\text{N} \sum\nolimits_{i=1}^N \left(\tilde{\boldsymbol{r}}_{i}\times{}^{\mathbb{C}}\boldsymbol{d}_{i}\right)\cdot\boldsymbol{n},
        \label{equ::average rotational slip}
\end{align}
where $\boldsymbol{S}_1$ represents the tendency of tangential slip, while $S_2$ describes the tendency of rotational slip. The value $N$ is the total number of the mark points. The vector $^\mathbb{C}d_{\mathrm{i}}$ denotes the displacement of mark point $i$ on the contact surface, where $\left\{\mathbb{C}\right\}$ denotes the reference frame of the tactile sensor. The vector $\tilde{\boldsymbol{r}}$ is the normalized vector from the mark point to the center of gravity, $\boldsymbol{n}$ is the normal vector of the contact surface.

\subsection{Action Formulation}

\begin{figure}[t]
        \centering
        \includegraphics[width = 0.9\linewidth]{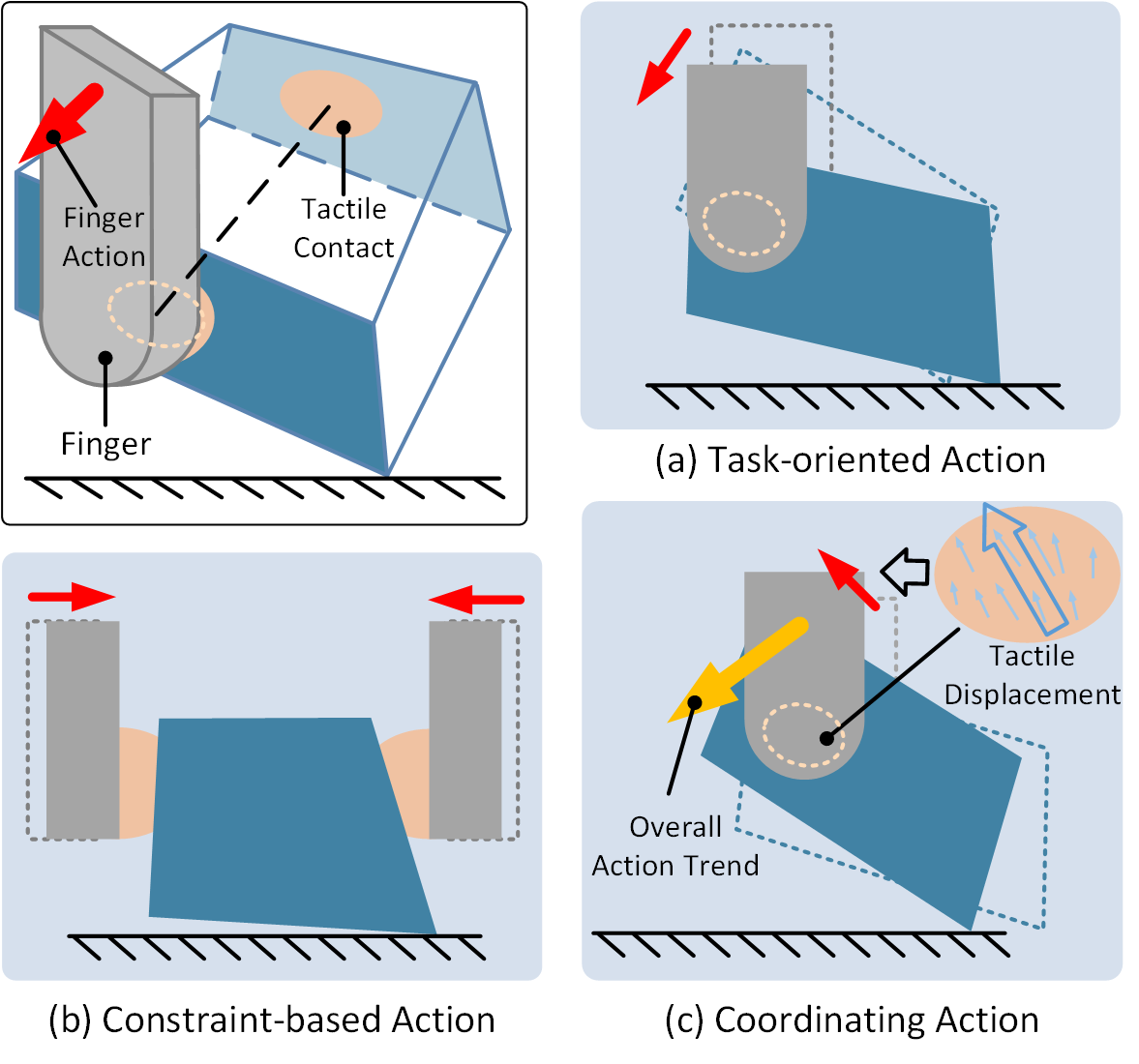}
        \caption{Reorientation with environment contact. The red arrows represent different actions.}
        \label{fig::rotating with environment}
\end{figure}

The form of the action depends on the reorientation scenario. 
First, we introduce the definitions of the various actions in the scenario where the object is in contact with the environment. The schematic diagrams illustrating these actions are provided in Fig. \ref{fig::rotating with environment}.

\textbf{Task-oriented Action.} The goal of task-oriented actions is to enable the pivot rotation of the object. The primary force driving the object's rotation comes from the environmental contacts. After the robot grasps the object, moving the hand closer to or away from the environment will modify the contact forces between the object and the environment. Therefore, the task-oriented action is defined as the displacement of the gripper during each control cycle:
\begin{equation}
        {}^{\mathbb{G}}\boldsymbol{v}_1=v_\mathrm{0}\boldsymbol{n}^\mathrm{task},
        \label{equ::task-oriented action}
\end{equation}
where $\left\{{\mathbb{G}}\right\}$ denotes the ground reference system, and $v_\mathrm{0}$ is the preset speed of the gripper's movement. The moving direction $\boldsymbol{n}^\mathrm{task}$ is set with an initial value, and will be further optimized online in the subsequent process.

\textbf{Constraint-based Action.} The purpose of the constraint-based action is to prevent the object from slipping out of the gripper. The force exerted by the environment on the object might increases the tangential force at the contact surface, thereby increasing the tendency to slide. Conversely, the grasping force applied by the fingers increases the normal force at the contact surface, which reduces the object's tendency to slide. To achieve the goal of the constraint-based action, there are two main strategies: reducing the contact force between the object and the environment or increasing the grasping force. Since reducing the contact force diminishes the tendency for rotation, increasing the grasping force is chosen as the constraint-based action:

\begin{equation}
        F_{t_n}=\begin{cases}F_{t_{n-1}}+\Delta F & \mathrm{if~}\|\boldsymbol{S}_1\|>{d}_\mathrm{lim}, \\
                F_{t_{n-1}} & \mathrm{else},
        \end{cases}
        \label{equ::constraint-based action}
\end{equation}
where $F_{t_n}$ denotes the grasping force of the $n$-th control cycle, and $\Delta F$ denotes the preset grasping force increment. Too much gripping force will reduce the tendency of the object to rotate, so the increase in grasping force is controlled using a predefined threshold ${d}_\mathrm{lim}$.

\textbf{Coordinating Action.} The coordinating action aims to resolve the conflict between the task-based action and the constraint-based action. If the object are continuously moving against the environment according to Equ. (\ref{equ::task-oriented action}), the tangential force required by the object may be further increased, thereby causing the object to slip, which violates the contact constraint condition of avoiding tangential slip. Therefore, new actions are needed to coordinate the above contradictions. In this work, a direct approach is adopted: the gripper moves in the direction of the tangential slip on the contact surface. First, the tangential slip direction is calculated using Equ. (\ref{equ::average tangential slip}). Then, the normalized vector in the tangential direction is obtained as
\begin{equation}
        ^\mathbb{C}\boldsymbol{e}_\mathrm{tan}=\frac{\boldsymbol{S}_1-\left(\boldsymbol{S}_1 \cdot \boldsymbol{n}\right) {}\boldsymbol{S}_1}{\|{}\boldsymbol{S}_1-\left(\boldsymbol{S}_1 \cdot \boldsymbol{n}\right) {}\boldsymbol{S}_1\|}
\end{equation}
where $n$ is the normal vector of the contact surface between the finger and the object, which can be calculated the tactile sensor. Additionally, when the contact surface is curved, the normal vectors at different points on the surface are computed, and the average normal is used.

Finally, the above normalized vector is used to determine the motion of the gripper:
\begin{equation}
        ^{\mathbb{G}}\boldsymbol{v}_2={}_{\mathbb{C}}^{\mathbb{G}}R{}^\mathbb{C}\boldsymbol{e}_\mathrm{tan}v_0={}_{\mathbb{H}}^{\mathbb{G}}R{}_{\mathbb{C}}^{\mathbb{H}}R{}^\mathbb{C}\boldsymbol{e}_\mathrm{tan}v_0,
        \label{equ::coordinating action}
\end{equation}
where ${}_{\mathbb{H}}^{\mathbb{G}}R$ represents the rotation matrix from the gripper to the ground, and ${}_{\mathbb{C}}^{\mathbb{H}}R$ represents the rotation matrix from the tactile sensor to the gripper. $v_\mathrm{0}$ is the preset speed of the gripper's movement.

The Equ. (\ref{equ::task-oriented action})--(\ref{equ::coordinating action}) define the three types of actions when there is contact with the environment. Similarly, the forms of actions in the absence of environmental contact can be derived and are provided in the Appendix.

\subsection{Action Optimization}
There are two action to be optimized. The first is the movement direction $\boldsymbol{n}^\mathrm{task}$ of the gripper in the scenario of contact with the environment, as defined in Equ. (\ref{equ::task-oriented action}). Another variable is the gripper rotation vector $\boldsymbol{n}^\mathrm{coor}$ in Equ. (\ref{equ::rotation in air}) in the scenario of not contact with the environment.

In order to avoid adding extra pre-train process and enable object reorientation when the object is first encountered, an online gradient optimization algorithm is applied. This algorithm determines the local gradient distribution using tactile feedback and then plans the next action to minimize the loss function. 

First, the loss function needs to be defined. There are two action goals in the reorientation task: (1) inducing object rotation, and (2) preventing macroscopic tangential slip on the contact surface. Therefore, the loss function can be determined as
\begin{equation}
        L(\boldsymbol{q}) = L_1 + L_2, \boldsymbol{q} = \boldsymbol{n}^\mathrm{task}, \boldsymbol{n}^\mathrm{coor}
        \label{equ::loss function}
\end{equation}
\begin{equation}
        L_1 = \|\boldsymbol{S}_1\|^2, 
        L_2 = \begin{cases}(\lambda_0 - S_2)^2 - \lambda_0^2 & \mathrm{if~} S_2>0, \\
                S_2^2 & \mathrm{else}.
                \end{cases}
        \label{equ::loss functions}
\end{equation}
where $\boldsymbol{q}$ denotes the vector to be optimized. The function $L_1$ is the loss function term that suppresses the macroscopic tangential slip of the contact surface, and $L_2$ is the loss function term that promotes the pivot rotation of the object. The value $\lambda$ is a preset value, which is used to ensure that when $S_2$ is greater than 0, $L_2$ also rises as $S_2$ rises.

Since there is no explicit relationship between the loss function and the vector $\boldsymbol{q}$ to be optimized, it is impossible to directly obtain the explicit form of the gradient. To address this problem, the finite difference method is used to approximate the gradient:

\begin{equation}
        \begin{bmatrix}\nabla_{\boldsymbol{q}} L(\boldsymbol{q})\end{bmatrix}_m\approx\frac{L(\boldsymbol{q}+\epsilon \boldsymbol{e}_m)-L(\boldsymbol{q}-\epsilon \boldsymbol{e}_m)}{2\epsilon}, m=1,2,3
        \label{equ::loss function gradient}
\end{equation}
where $\boldsymbol{e}_m$ is the unit vector in the $m$-th direction and $\epsilon$ is a small perturbation value. The above equation shows that the gripper needs to try to change a component $\boldsymbol{e}_m$ of the vector to be optimized, and determine the gradient of the loss function on the three degrees of freedom according to the feedback.

After completing the gradient update, the vector $\boldsymbol{q}$ to be optimized is updated with a learning rate $\alpha$ using the following equation:
\begin{equation}
        \boldsymbol{q}^{(k+1)}=\boldsymbol{q}^{(k)}-\alpha\nabla_{\boldsymbol{q}} L(\boldsymbol{q})
\end{equation}

To acquire the gradient of the loss function, the gripper needs to move sequentially in different directions $\boldsymbol{e}_m$. To prevent the gradient update from interfering with other sub-actions, the following strategies are implemented: first, the task-based action is executed before the other two actions; second, after exploring along all the $\boldsymbol{e}_m$, the gripper's position is ensured to be exactly the same as the position after moving according to Equ. (\ref{equ::task-oriented action}) (for optimizing $\boldsymbol{n}^\mathrm{task}$). Similarly, after rotating along three axes, the posture is kept consistent with the original posture after rotating according to Equ. (\ref{equ::rotation in air}) (for optimizing $\boldsymbol{n}_{\mathrm{rot}}$).

\section{Experiment}


Two sets of experiments were conducted: an ablation study to evaluate the impact of different actions and the online optimization method on success rates, and a demonstration to assess the method's performance across various objects, reorientation scenarios, and environmental constraints. In the experiments, a visual system was used to measure the object's orientation in real-time, serving as the control signal to determine whether the robot should continue its motion. The tactile sensing was performed using a Tac3D sensor, which can measure real-time contact deformation at a frequency exceeding 30 Hz \cite{zhang2024evaluation}.

\subsection{Ablation Models}

The experiment defines successful object reorientation as achieving a target orientation with a deviation of less than 5 degrees. 
The experimental results revealed two primary failure modes: (1) excessive tangential slippage at the contact surface, exceeding 2 cm, and (2) failure to achieve object reorientation. These are labeled as SL (slipped) and ST (stopped), respectively.

\begin{figure}[t]
        \centering
        \includegraphics[width = 1\linewidth]{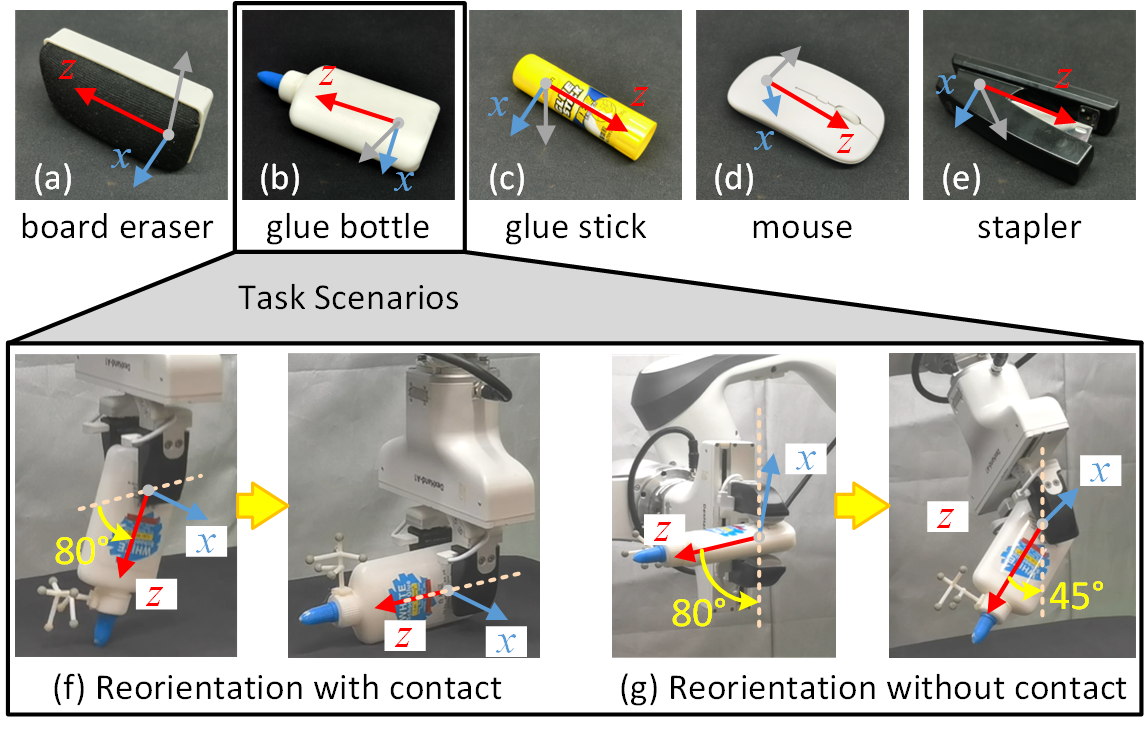}
        \caption{Objects (a)--(e) and their respective coordinate frames are shown. (f) is the initial and target states of object reorientation with environmental contact, while (g) shows the initial and target states of object reorientation without environmental contact.}
        \label{fig::ablation experiment}
\end{figure}

\textcolor{g}{As shown in Fig. \ref{fig::ablation experiment}, the tests were conducted using five everyday objects: the board eraser, representing a soft object; the glue bottle, representing an object with a shifting center of mass; the glue stick, representing a textured object; the mouse, representing a curved object; and the stapler, representing an asymmetric object.} Ablation experiments were conducted under two contact conditions: one without environmental contact and one with environmental contact.

\begin{figure}[t]
        \centering
        \includegraphics[width = 1.0\linewidth]{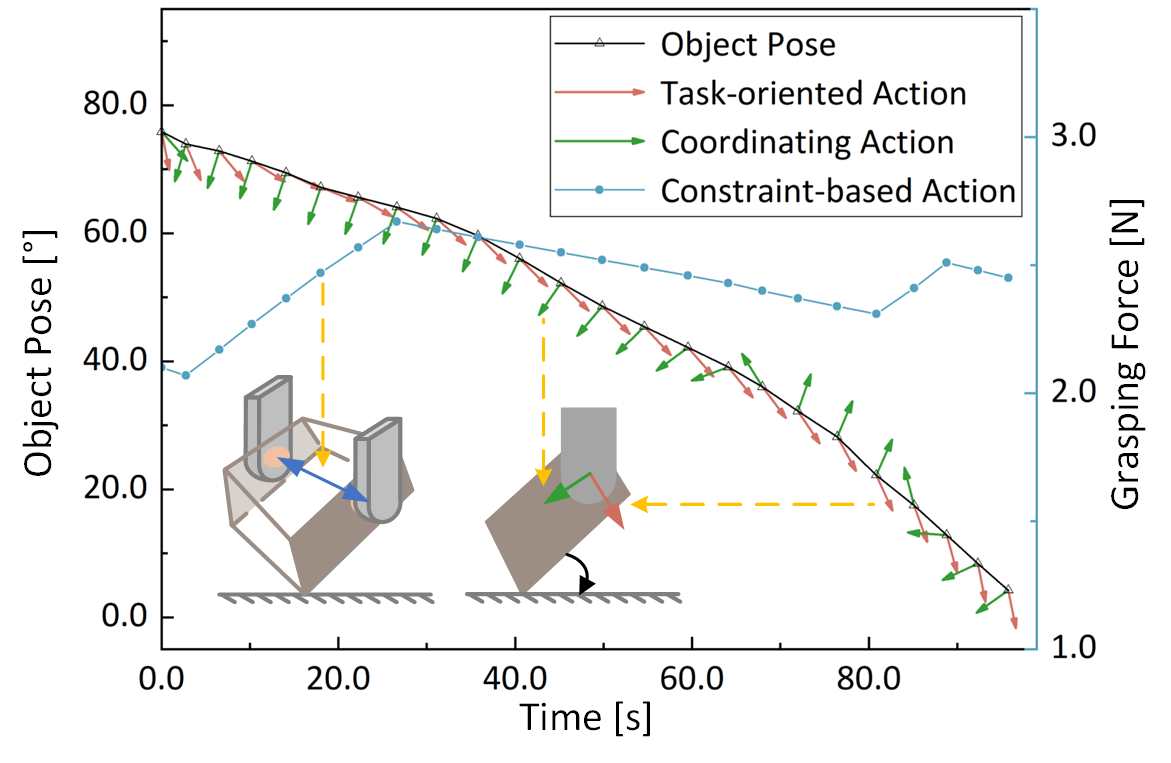}
        \caption{\textcolor{g}{Results for the glue bottle. The black line represents the angle between the object's posture and the target posture. The task-oriented actions (red arrows) and coordinating actions (green arrows) represent the movements of the gripper, while the constraint-based action (blue line) shows the change in grasping force. The three types of actions are performed sequentially at different time points.}}
        \label{fig::results}
\end{figure}

Each contact condition was divided into four experimental groups: (1) NTO: No Task-Oriented Action; (2) NCB: No Constraint-Based Action; (3) NC: No Coordinating Action; and (4) NOA: No Online Adjustment. Additionally, a control group utilizing a three-action superposition and online optimization method was established, abbreviated as Control Group (CG). The results of these experiments are summarized in Fig. \ref{fig::results}, Tab. \ref{tab:results with object-environment contact} and Tab. \ref{tab:results without object-environment contact}.

When employing all three action superpositions along with the online optimization algorithm, the method successfully completed the task across different objects and scenarios. In scenarios without environmental contact, the experimental group without constraint-based actions (NCB) also achieved successful object reorientation (the second column of table \ref{tab:results without object-environment contact}). A key reason for this is that the gravity acting on an eccentrically grasped object in mid-air tends to induce rotation rather than macroscopic tangential slippage. However, when the object is in contact with the environment, the contact forces counteract this rotational tendency caused by gravity, resulting in a lower success rate (as shown in the second column of table \ref{tab:results with object-environment contact}).

\begin{table}[t]
        \centering
        \caption{Reorientation Results with Object-Environment Contact}
        \label{tab:results with object-environment contact}  
        \begin{tabular}{cccccc}
        \toprule	
        \makebox[0.15\linewidth][c]{Objects} & 
        \makebox[0.1\linewidth][c]{NTO} &
        \makebox[0.1\linewidth][c]{NCB} & 
        \makebox[0.1\linewidth][c]{NC} & 
        \makebox[0.1\linewidth][c]{NOA} &
        \makebox[0.1\linewidth][c]{CG}\\
        \midrule
        board eraser & ST  & \checkmark & SL & ST\&SL & \checkmark\\
        glue bottle  & ST  & \checkmark & SL & ST\&SL & \checkmark\\
        glue stick   & ST  & SL         & SL & ST\&SL & \checkmark\\
        mouse        & SL  & SL         & SL & ST\&SL & \checkmark\\
        stapler      & ST  & \checkmark & \checkmark & SL & \checkmark\\
        \bottomrule
        \end{tabular}
        \begin{tablenotes}
		\item NTO: No Task-Oriented Action;  NCB: No Constraint-Based Action
		\item NC: No Coordinating Action;\quad\, NOA: No Online Adjustment.
		\item CG: Control Group;\quad\quad ST: Stopped ;\quad\quad SL: Slipped
        \end{tablenotes}
\end{table}

\begin{table}[t]
        \centering
        \caption{Reorientation Results without Object-Environment Contact}
        \label{tab:results without object-environment contact}  
        \begin{tabular}{cccccc}
        \toprule	
        \makebox[0.15\linewidth][c]{Objects} & 
        \makebox[0.1\linewidth][c]{NTO} &
        \makebox[0.1\linewidth][c]{NCB} & 
        \makebox[0.1\linewidth][c]{NC} & 
        \makebox[0.1\linewidth][c]{NOA} &
        \makebox[0.1\linewidth][c]{CG}\\
        \midrule
        board eraser & ST  & \checkmark & SL & SL & \checkmark\\
        glue bottle  & \checkmark  & \checkmark & SL & ST & \checkmark\\
        glue stick   & \checkmark  & \checkmark & SL & SL & \checkmark\\
        mouse        & ST  & \checkmark & SL & ST & \checkmark\\
        stapler      & \checkmark  & \checkmark & ST\&SL & ST & \checkmark\\
        \bottomrule
        \end{tabular}
\end{table}

The experimental results indicate that task-oriented actions aim to promote object rotation, and disabling these actions prevents the object from rotating. Constraint-based actions are responsible for minimizing macroscopic slippage at the contact surface, and disabling these actions leads to excessive slippage. Coordinating actions work to balance the task- and constraint-based actions, and disabling them may result in either failure type. Finally, the role of the online optimization method is to allow the robot to explore and identify motion directions that satisfy both task and constraint requirements during object reorientation; therefore, omitting the optimization process increases the likelihood of both failure types.

\subsection{Demonstration of Charger Reorientation Task}
Next, the two contact scenarios were applied to a typical everyday in-hand reorientation task: inserting a charger into a charging pad. The process involved the following steps: (1) Grasping the charger from the ground; (2) Utilizing a no-contact reorientation strategy to rotate the charger until it is aligned parallel to the fingers; (3) Employing a contact-based reorientation strategy for further adjustment of the charger's orientation; and (4) Fine-tuning the robotic arm's pose to align the charger with the charging pad and complete the connection.

\begin{figure}[t]
        \centering
        \includegraphics[width = 1\linewidth]{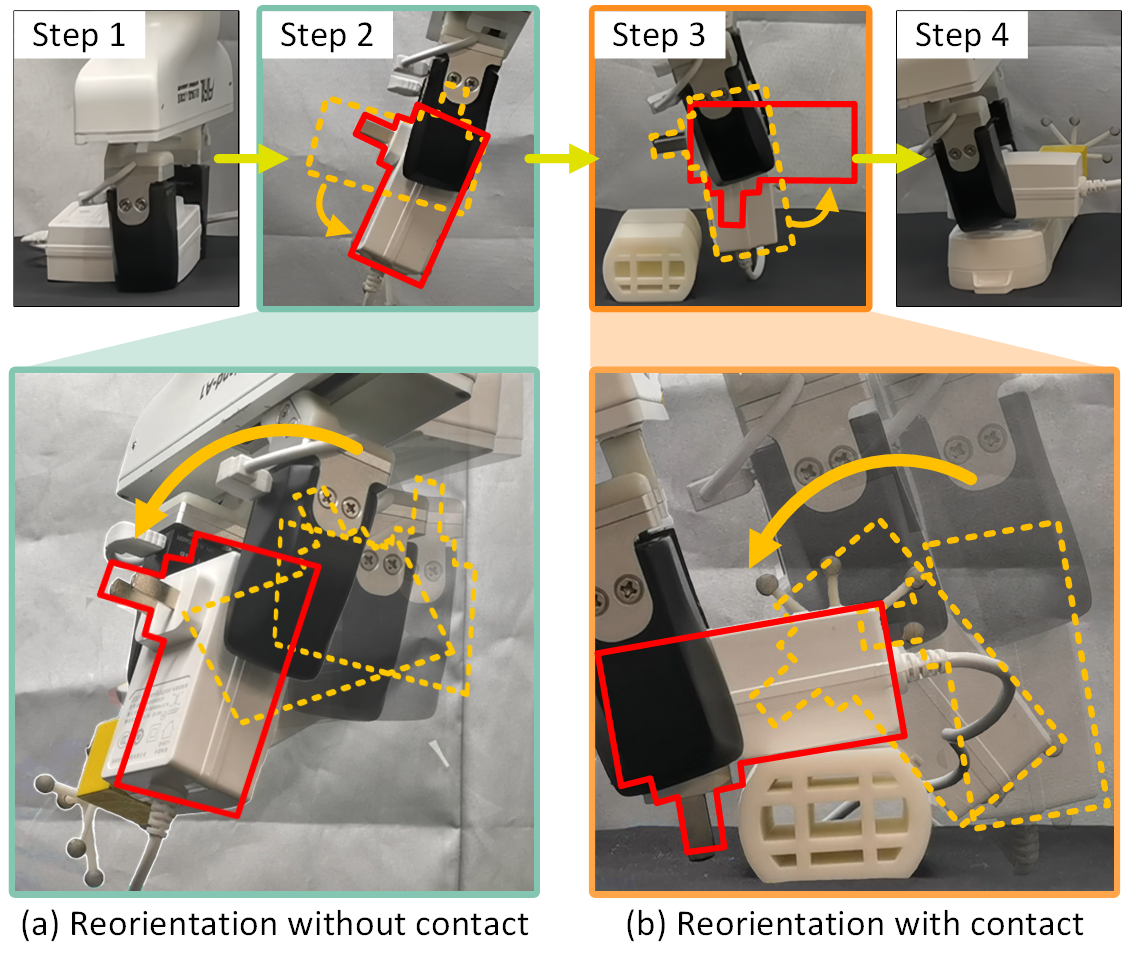}
        \caption{The process of reorient the charger. First reorient the object without environment contact, then reorient the object with environment contact. The shape and position of the obstacle are unknown.}
        \label{fig::Demonstration experiment}
\end{figure}

In this experiment, adjusting the robotic arm was necessary to align the charger with the charging pad. However, arm pose adjustments alone were insufficient, as the charger was oriented towards the palm, necessitating an in-hand reorientation that could not be achieved through arm pose adjustments alone. To verify the desired orientation, objects were equipped with target markers, enabling control over transitions between different reorientation strategies. The robotic arm's motion planning relied solely on real-time tactile feedback.


In the experiment, the object had an unknown shape and position, and was not fixed to the table, increasing environmental uncertainty. The charging cable also applied continuous disturbances, further complicating the task. Despite these challenges, the final results, as illustrated in Fig.\ref{fig::Demonstration experiment}, show that successful object reorientation was achieved across different scenarios, even in the presence of unknown and dynamic environments as well as external disturbances.
 
\section{CONCLUSIONS}
\textcolor{g}{In this paper, we propose a restructured approach to the object reorientation problem by decomposing the task into three types of actions and employing an online optimization method. Unlike modeling-based approaches, our method eliminates the need for prior knowledge of object shapes or contact models. Additionally, it can be directly deployed in novel scenarios without requiring pre-training. Specifically, our approach simplifies input by reducing the perception dimension to desired tactile feedback, and it categorizes actions to reduce the complexity of controlling multi-degree-of-freedom systems. Experimental results demonstrate the effectiveness of the framework in both contact and non-contact reorientation scenarios, with robust adaptation to varying environmental constraints.}

In this paper, actions were categorized into three types, with their forms predefined. In future work, we plan to integrate learning-based methods, enabling the robot to autonomously discover optimal action forms. By doing so, we aim to extend the framework to multi-fingered grippers and evaluate its adaptability to other tasks, such as tool use.

\addtolength{\textheight}{-12cm}   

\section*{APPENDIX: Reorientation in the Air}

\textbf{Task-oriented Action.} When the grasping position is offset from the center of gravity, gravitational torque aids the rotation caused by the fingers. To enhance pivot rotation, reducing the torque from the fingers is needed. Thus, the task-oriented action is designed to lower the gripping force:

\begin{equation}
        F_{t_n}=\begin{cases} F_{t_{n-1}}-\Delta F & \mathrm{if~}\|{}\boldsymbol{S}_1\|<{d}_\mathrm{lim}\\
        F_{t_{n-1}} & \mathrm{ else}\end{cases}
\end{equation}

To prevent tangential sliding caused by a rapid reduction in grasping force, the force is decreased only when the average displacement of the contact surface is below a set threshold.

\textbf{Constraint-based Action.} To prevent slippage on the contact surface, a straightforward approach is to increase the gripping force, thereby enhancing the surface's ability to resist tangential forces. To avoid applying excessive gripping force, increase the force only when you detect a tendency for the contact surface to slip tangentially:

\begin{equation}
        F_{t_n}=\begin{cases}F_{t_{n-1}}+\Delta F & \mathrm{if~}\|{}\boldsymbol{S}_1\|>{d}_\mathrm{lim}, \\
        F_{t_{n-1}} & \mathrm{else}
        \end{cases}
        \label{equ::constraint-based action air}
\end{equation}

\textbf{coordinating Action.} Object reorientation requires reducing the grasping force, while preventing slippage calls for increasing it. Coordinating actions address this by adjusting the gripper's posture to enhance rotation while minimizing slip risk. Changing the object's posture can be effective, as it alters the gravitational torque on the object. Changing the manipulator's orientation is defined as an RPY (roll-pitch-yaw) rotation:
\begin{equation}
        \boldsymbol{{R}}({\theta_1,\theta_2,\theta_3}) = \boldsymbol{R}(^\mathbb{G}\boldsymbol{Z},\theta_1)\boldsymbol{R}(^\mathbb{G}\boldsymbol{Y},\theta_2)\boldsymbol{R}(^\mathbb{G}\boldsymbol{X},\theta_3)
        \label{equ::sequences rotation in air}
\end{equation}

\begin{equation}
        \boldsymbol{n}^\mathrm{coor} = [\theta_1, \theta_2, \theta_3]^\mathrm{T}
        \label{equ::rotation in air}
\end{equation}

The rotation angle is limited to 3$^\circ$ in each control cycle. Other algorithms will be used to further optimize the rotation direction $\boldsymbol{n}^\mathrm{coor}$ (see Section \uppercase\expandafter{\romannumeral3}-D for details).

\addtolength{\textheight}{+12cm} 

\normalem
\bibliographystyle{IEEEtran}
\bibliography{ref}

\end{document}